\documentclass[conference]{IEEEtran}
\ifCLASSINFOpdf
\else
\fi

\usepackage{amsmath}
\usepackage{bm}
\usepackage{mathrsfs}
\usepackage{graphicx}
\usepackage{algorithm}
\usepackage{algorithmic}
\usepackage{multirow}
\usepackage{booktabs}
\usepackage[dvipsnames]{xcolor}
\usepackage{amsfonts}
\usepackage{url}

\begin{document}
%
\title{Benchmarking Transferable Adversarial Attacks}



%
\author{\IEEEauthorblockN{Zhibo Jin\IEEEauthorrefmark{1},
Jiayu Zhang\IEEEauthorrefmark{2},
Zhiyu Zhu\IEEEauthorrefmark{1}and 
Huaming Chen\IEEEauthorrefmark{1}
}
\IEEEauthorblockA{\IEEEauthorrefmark{1}The University of Sydney}
\IEEEauthorblockA{\IEEEauthorrefmark{2}Suzhou Yierqi}
}


\IEEEoverridecommandlockouts
\makeatletter\def\@IEEEpubidpullup{6.5\baselineskip}\makeatother
\IEEEpubid{\parbox{\columnwidth}{
    Network and Distributed System Security (NDSS) Symposium 2024\\
    26 February - 1 March 2024, San Diego, CA, USA\\
    ISBN 1-891562-93-2\\
    https://dx.doi.org/10.14722/ndss.2024.23xxx\\
    www.ndss-symposium.org
}
\hspace{\columnsep}\makebox[\columnwidth]{}}

\maketitle

\begin{abstract}
The robustness of deep learning models against adversarial attacks remains a pivotal concern. 
This study presents, for the first time, an exhaustive review of the transferability aspect of adversarial attacks. 
It systematically categorizes and critically evaluates various methodologies developed to augment the transferability of adversarial attacks. This study encompasses a spectrum of techniques, including Generative Structure, Semantic Similarity, Gradient Editing, Target Modification, and Ensemble Approach. Concurrently, this paper introduces a benchmark framework \textit{TAA-Bench}, integrating ten leading methodologies for adversarial attack transferability, thereby providing a standardized and systematic platform for comparative analysis across diverse model architectures. Through comprehensive scrutiny, we delineate the efficacy and constraints of each method, shedding light on their underlying operational principles and practical utility. 
This review endeavors to be a quintessential resource for both scholars and practitioners in the field, charting the complex terrain of adversarial transferability and setting a foundation for future explorations in this vital sector. The associated codebase is accessible at: \url{https://github.com/KxPlaug/TAA-Bench}
\end{abstract}


%


\section{Introduction}
In recent years, adversarial attacks have emerged as a significant research direction for artificial intelligence and machine learning, especially in the context of the security of deep learning. It originates from the observation that deep neural networks (DNNs) are sensitive to subtle perturbations in input data. Even imperceptible to the human eye, such changes can lead to incorrect output results~\cite{goodfellow2014explaining}. Adversarial attacks can be categorized into two types based on the availability of model data: white-box attacks and black-box attacks~\cite{chakraborty2018adversarial, xu2020adversarial, li2022review, zhang2022investigating}. White-box attacks assume the model's internal information are accessible, such as its parameters, structure, and training data. In contrast, black-box attacks occur without knowledge of the internal information of the attacked model, which aligns more closely with real-world scenarios, as attackers often have constraints in access operation.

From the technical perspective, black-box attacks can be classified into two categories: query-based attacks and transferable adversarial attacks. The former, although unable to directly access the model's content, requires multiple requests to the target model~\cite{zhou2022adversarial}. Thus, attackers can infer and construct an approximate model to launch the attack. A limitation is that it necessitates regular and extensive access to the target model, thereby reducing the stealthiness of the attack. On the contrary, our emphasis is on a more practical form of attack, namely transferable adversarial attacks (TAA). This method involves the use of a surrogate model, which, while distinct from the target model, shares similar features or functionalities. Adversarial samples generated on this surrogate model are then used to attack the target model~\cite{zhao2021success}. Owing to certain generalization properties inherent in deep learning models, these adversarial samples often successfully mislead the target model, different from the surrogate, thus facilitating a transfer attack. The key advantage of this method lies in the ability to operate without direct access or querying of the target model, thereby enhancing the stealth and practicality of the attack. 

Despite the widespread research on transferable adversarial attacks in recent years, it lacks a comprehensive and systematic retrospective study. Thus, this paper aims to systematically review and categorise the existing classical and latest TAA methods. For the first time, we investigate TAA methods from multiple dimensions, categorising existing approaches into generative architecture, construction of semantic similarity, gradient editing types, modification of attack targets and ensemble types. Additionally, we benchmark baseline methods for comparison. Moreover, we reproduce 10 representative methods of transferable adversarial attacks and integrated these methods into an open-source benchmark framework, which is published on GitHub \textit{TAA-Bench}, facilitating related researches. Our main contributions are:
\begin{itemize}
    \item We thoroughly collate existing methods of transferable adversarial attacks, and systematically analyse their implementation principles.
    \item We present an extensible, modular and open-source benchmark \textit{TAA-Bench} that includes implementations of different types of transferable adversarial attacks to facilitate research and development in this field.
\end{itemize}


\section{Problem Definition}

In the study of transferability in adversarial attacks, our objective is to generate a slightly perturbed input sample \(x^\prime\), causing misclassification of a black-box target deep learning model while keeping these changes imperceptible to human observers. Specifically, consider a surrogate deep learning model \(f\) with parameters \(\theta\), a black-box target deep learning model \(f^\prime\), a representative input sample \(x\) along with its corresponding true label \(y\), and a small perturbation magnitude \(\epsilon\). Our goal is to find a perturbation \(\delta\) such that \(x^\prime = x + \delta\) satisfies two conditions: 1) the target model \(f^\prime\) does not output the true label \(y\) when predicting the input sample \(x^\prime\), i.e., \(f^\prime(x^\prime) \ne y\); 2) the magnitude of the perturbation \(\delta\) is constrained to be within a given threshold \(\epsilon\), ensuring that \(x^\prime\) remains indistinguishable from original sample \(x\) to human.

\section{Systemization of Knowledge}

\begin{table*}[htpb]
\centering
\caption{Comparative Analysis of Adversarial Attack Strategies}
\label{tab:Comp}
\resizebox{\textwidth}{!}{%
\begin{tabular}{|l|l|l|l|l|}
\hline
\textbf{Strategy} & \textbf{Description} & \textbf{Examples} & \textbf{Advantages} & \textbf{Disadvantages} \\ \hline
Generative Architecture & Generates perturbations via a dedicated network & \cite{xiao2018generating,zhu2024geadvgan} & Fast attack execution & Requires extra network training, complex training process \\ \hline
Semantic Similarity & Uses semantically similar samples in attacks & \cite{xie2019improving,lin2019nesterov,long2022frequency,wu2023towards,wang2023improving,zhu2023improving} & Simple to deploy and motivate & High-quality sample generation can be challenging \\ \hline
Gradient Editing & Aims to reduce gradient overfitting & \cite{dong2018boosting,zhang2023transferable,zhu2024geadvgan} & Independent of target attack characteristics & Can reduce attack accuracy, limited transferability \\ \hline
Target Modification & Focuses on common features in different models & \cite{wang2021feature,zhang2022improving,jin2023danaa,ma2023transferable} & Exploits model similarities & Relies on model feature commonalities, difficult to implement \\ \hline
Ensemble Approach & Integrates multiple models for attacks & \cite{liu2016delving,xiong2022stochastic} & Enhances transferability and robustness & More computationally intensive, requires multiple model integration \\ \hline
\end{tabular}%
}
\end{table*}

In this section, we provide a comprehensive summarisation to a variety of existing transferable attack methods, each enhancing the transferability of adversarial attacks from different perspectives. The first category is for baseline methods, which, while not specifically optimised for attack transferability, utilize classic white-box attack methods to assess the effectiveness of other transferable methods. As depicted in Table~\ref{tab:Comp}, we categorise the transferable attacks into five types: Generative Architecture, Semantic Similarity, Gradient Editing, Target Modification, and Ensemble Approach.

\subsection{Baseline approaches}

To study transferable adversarial attacks, the selection of a suitable baseline method is essential for benchmarking attack techniques. In this work, we select the Iterative Fast Gradient Sign Method (I-FGSM) as the baseline~\cite{dong2018boosting,xie2019improving}. As an enhancement of FGSM, I-FGSM applies iterative refinements to generate more effective adversarial samples. Its lack of specific optimisation for transferability provides unified context for assessing other methods. Superior performance over I-FGSM in transferability is indicative of enhanced attack efficacy.

The principle of I-FGSM involves iterative applications of the Fast Gradient Sign Method. Starting with an input \( x_0 \), each iteration computes the loss function gradient \( L \) relative to the current sample \( x_i \), aiming to maximize loss and induce misclassification. Adjustments are based on the gradient sign, following \( x_{i+1} = x_i + \epsilon \cdot \operatorname{sign}(\nabla_x L(\theta, x_i, y)) \), where \( \epsilon \) controls perturbation magnitude, \( \theta \) denotes model parameters, and \( y \) the true label. The update size is consistent across pixels, guided by the gradient direction. The process concludes after a predetermined number of iterations or upon achieving misclassification.

\subsection{Generative Architecture}
This category of methods employs Generative Adversarial Networks (GANs) to produce adversarial samples. The core idea of these methods lies in using generative models to mimic the decision boundaries of the target attack models, thereby generating efficient transferable adversarial samples. The advantage of such methods is that once the Generator is trained, adversarial samples can be quickly generated without further querying the model. Examples of this type of attack include AdvGAN~\cite{xiao2018generating} and GE-AdvGAN~\cite{zhu2024geadvgan}.

\subsubsection{AdvGAN~\cite{xiao2018generating}}

The main principle of AdvGAN is based on Generative Adversarial Networks (GANs), which include a generator and a discriminator. The generator creates slight perturbations and adds them to the original input data, generating counterfeit samples. The discriminator learns to distinguish between fake and real samples, which are then assessed by the target neural network to evaluate the classification effect of the perturbed samples. The adversarial samples generated by AdvGAN aim to deceive the target network into making incorrect classifications while remaining imperceptible to human observers. This method effectively combines the generative capabilities of GANs with the requirements of adversarial attacks, functioning efficiently in both semi-white-box and black-box attack scenarios.

The overall objective of AdvGAN is to find a balance between generating adversarial samples and deceiving the model. Thus, the total loss is the sum of the discriminator loss and the generator's impact on the target model: \small \[ \operatorname{min}_G \operatorname{max}_D \mathbb{E}_{x, y \sim \text{data}} [\log D(x) + \log(1 - D(G(x))) + \lambda L(f(G(x)), y)] \], where \(\lambda\) is a weight coefficient used to balance the two objectives. Through this method, AdvGAN can generate adversarial samples that are as similar to real samples as possible but can mislead the target model.

\subsubsection{GE-AdvGAN~\cite{zhu2024geadvgan}}
GE-AdvGAN, compared to AdvGAN, has been optimized in terms of transferability and has also improved the efficiency of the algorithm. The core idea is the optimization of the gradient update method during the generator training process. GE-AdvGAN introduces a novel Gradient Editing (GE) mechanism, utilizing frequency domain exploration to determine the direction of gradient editing. This method enables GE-AdvGAN to generate highly transferable adversarial samples while significantly reducing the execution time to generate these samples.

Specifically, in AdvGAN, the Generator's loss is divided into three components: \(L_{adv}\), \(L_{GAN}\), and \(L_{hinge}\). These respectively represent the loss functions for the attack, generation, and control of perturbations. The portion controlling the attack can be decomposed as \(\nabla \theta_G L_{adv} = \frac{\partial L_{adv}}{\partial (x+G(x))} \cdot \frac{\partial (x+G(x))}{\partial G(x)} \cdot \frac{\partial G(x)}{\partial \theta_G}\). In GE-AdvGAN, the term \(\frac{\partial (x+G(x))}{\partial G(x)}\) is replaced with \(-\operatorname{sign} \left( \frac{1}{N}\sum_{i=1}^{N}\frac{\partial L(x_{f_i},y)}{\partial x_{f_i}}   \right)\), where \(x_{f_i}\) are samples generated using frequency domain exploration.


\subsection{Semantic Similarity}
The core concept of this category of methods is to find a sample and construct samples that are semantically related to it, and simultaneously attack these semantically related samples, thereby expanding the transferability of adversarial attacks. Such attack methods are represented by Diverse Input Fast Gradient Sign Method (DI-FGSM)~\cite{xie2019improving}, Scale-Invariant Nesterov Iterative Fast Gradient Sign Method (SI-NI-FGSM)~\cite{lin2019nesterov}, Spectrum Simulation Attack (SSA)~\cite{long2022frequency}, Centralized Perturbation Attack (CPA)~\cite{wu2023towards}, Feature Disruptive Universal Adversarial Attack (FDUAA)~\cite{wang2023improving}, and Structure Invariant Attack (SIA)~\cite{wang2023structure}.

\subsubsection{DI-FGSM~\cite{xie2019improving}}

The core principle of the DI-FGSM is to introduce input diversity in the process of generating adversarial samples to find Semantic Similarity. This is achieved by applying random transformations (such as resizing and padding) to the input image in each iteration. These variations create diverse input patterns, helping to prevent overfitting to specific network parameters, thereby enhancing the effectiveness of the generated adversarial samples against different models.

Assume the original input image is \(x\), and the adversarial sample is initialized as \(x_{0}^{\prime} = x\). For each iteration \(i\), a random transformation \(\tau\) is applied to the current adversarial sample \(x_{i}^{\prime}\), resulting in the transformed sample \(\widetilde{x}_{i}^{\prime} = \tau(x_{i}^{\prime})\). Then, the gradient of the loss function \(L(f(\widetilde{x}_{i}^{\prime}), y)\) with respect to \(\widetilde{x}_{i}^{\prime}\) is computed, where \(y\) is the target label. The adversarial sample is updated using this gradient:
\small \[ x_{i+1}^{\prime} = x_{i}^{\prime} + \epsilon \cdot \operatorname{sign} (\nabla_{x_{i}^{\prime}}L(f(\widetilde{x}_{i}^{\prime}), y)) \]
Here, \(\epsilon\) is the step size, and the sign function returns the sign of the gradient. By repeating this process, DI-FGSM increases the transferability of the adversarial samples, making them more likely to be effective on unknown models.

\subsubsection{SI-NI-FGSM~\cite{lin2019nesterov}}

The SI-NI-FGSM is an adversarial attack algorithm that integrates Scale Invariance (SIM) and the Nesterov Iterative Method (NIM). This method enhances the effectiveness and transferability of adversarial samples by introducing NIM and SIM on top of the Fast Gradient Sign Method (FGSM). SI-NI-FGSM first utilizes NIM to predict future changes in the gradient for more precise updates of the adversarial samples. It then maintains scale invariance by adjusting the scale of the input image, thus improving the transferability of the attack across different models.

Specifically, SI-NI-FGSM initially pre-updates the input sample using the Nesterov method, calculated with the formula \(x^\prime = x + \alpha \cdot v\), where \(x\) is the current sample, \(v\) is the accumulated gradient, and \(\alpha\) is the pre-update step length. It then calculates the gradient at the pre-update point \(g = \nabla_x L(\theta, x^\prime, y)\) and updates the momentum \(v = \mu \cdot v + g\). Finally, the sample is updated using \(x = x + \epsilon \cdot \operatorname{sign}(v)\). During the generation of adversarial samples, the scale of the input image is adjusted to ensure that the generated perturbation maintains the same effect on images of different scales.

\subsubsection{SSA~\cite{long2022frequency}}

The core principle of SSA is to simulate diverse models in the frequency domain, thereby enhancing the transferability of samples. The specific operation includes using DCT and inverse DCT to transform the spectral signature of the input image, generating diverse spectral saliency maps, which indicate the diversity of substitute models. The approach further includes randomly masking features in the frequency domain to identify and exploit similar semantics, thereby accomplishing transferability in the attack.

Specifically, SSA first uses DCT to transform the input image from the spatial domain to the frequency domain. This process can be mathematically represented as \(\mathcal{D}(\bm{x}) = \bm{A}\bm{x}\bm{A}^\mathrm{T}\), where \(\bm{A}\) is an orthogonal matrix. Subsequently, SSA introduces a Spectrum Saliency Map, defined as the response of the spectrum of the input image to the gradient of the model's loss function, expressed as \small \(\bm{S}_\phi = \frac{\partial L(\mathcal{D_I}(\mathcal{D}(\bm{x}), y; \phi)}{\partial \mathcal{D}(\bm{x})}\). Here \(\mathcal{D_I}\) is the inverse DCT transform, used to convert frequency domain data back to the spatial domain. Finally, SSA employs a random spectral transformation \(\mathcal{T}(\cdot)\), which can be expressed as \(\mathcal{T}(\bm{x}) = \mathcal{D_I}(\mathcal{D}(\bm{x}) + \mathcal{D}(\bm{\xi}) \odot \bm{M})\), where \(\odot\) denotes the Hadamard product, and \(\xi\) and \(\bm{M}\) are variables randomly sampled from Gaussian and uniform distributions, respectively. This transformation produces diverse spectral saliency maps, thereby simulating different substitute models and enhancing the transferability of adversarial samples.

\subsubsection{CPA~\cite{wu2023towards}}

The principle of CPA~\cite{wu2023towards} is to enhance the transferability of adversarial attacks through precise perturbation optimization in the frequency domain on DNNs. This method first employs DCT to decompose data into the frequency domain, thereby facilitating the exploration of similar semantics. Then, it reduces unnecessary perturbations by quantizing each Y/Cb/Cr channel and focuses the optimization on the main frequency coefficients that influence model predictions. Finally, the differential quantization matrix is optimized through backpropagation, ensuring that perturbations are concentrated in the dominant frequency areas. The key to this method lies in effectively centralizing and optimizing perturbations, thereby improving the transferability of adversarial samples and their ability to bypass defense mechanisms.

\subsubsection{FDUAA~\cite{wang2023improving}}
The core principle of the FDUAA is to generate universally transferable adversarial perturbations (UAPs) by disrupting features that are not dependent on specific model architectures, such as edges or simple textures. Specifically, this method weakens important channel features while enhancing less significant ones, as determined by a specific strategy, through a target function. Additionally, the method iteratively updates UAPs using the average gradient of small-batch inputs to capture local information. It also introduces a momentum term to accumulate gradient information from iterative steps, sensing the global information of the entire training set.

\subsubsection{SIA~\cite{wang2023structure}}

The principle of the SIA is based on applying a series of random transformations to an image, aiming to create diverse adversarial samples with structural characteristics. The SIA method processes the image in blocks, applying random image transformations such as rotation and scaling to each block, thereby increasing the diversity of the samples and finding similar semantics. This method maintains the basic structure of the original image while generating challenging adversarial samples capable of effectively deceiving deep neural networks. The key to SIA lies in its ability to enhance the transferability of the samples by introducing transformations, while simultaneously maintaining the structural integrity of the image.

\subsubsection{FSPS~\cite{zhu2023improving}}
The FSPS method is a novel algorithm designed to enhance the transferability of adversarial attacks in machine learning. This approach is centered around two fundamental concepts: identifying stationary points on a loss curve and executing frequency-based searches from these identified points. The process is initiated by pinpointing stationary points on the loss curve, defined as locations where the derivative of the loss function is zero. These identified points serve as the starting points for the attack. Subsequently, FSPS applies a frequency-based search methodology to scrutinize the most effective adversarial directions in the vicinity of these stationary points.

\subsection{Gradient Editing}

This category of methods focuses on modifying or optimizing gradient information to generate adversarial samples. These techniques often rely on a deep understanding and manipulation of gradients in surrogate models, to make the generated samples effective on target models. Representative methods include Momentum Iterative Fast Gradient Sign Method (MI-FGSM)~\cite{dong2018boosting}, Token Gradient Regularization (TGR)~\cite{zhang2023transferable}, Frequency-based Stationary Point Search (FSPS)~\cite{zhu2023improving}, and the previously mentioned GE-AdvGAN~\cite{zhu2024geadvgan}. Since GE-AdvGAN has already been discussed earlier, it will not be elaborated upon in this section.

\subsubsection{MI-FGSM~\cite{dong2018boosting}}

The MI-FGSM integrates a momentum term in its iterative process to stabilize the update direction and escape from local maxima, thereby generating more transferable adversarial samples. In each iteration, it accumulates a velocity vector in the direction of the loss function gradient, aiding in optimizing stability and avoiding suboptimal local maxima.

In MI-FGSM, an adversarial perturbation rate \(\alpha\) is set, proportional to the total perturbation limit \(\epsilon\) and the number of iterations \(T\). The method starts with the original input \(x\) and initializes a zero vector \(g\) as the starting value for momentum. In each iteration, it first calculates the gradient of the loss function \(\nabla_xL(x_t, y)\) for the current adversarial sample \(x_t\), then combines this gradient with the previous momentum \(g_t\), weighted by the momentum factor \(\mu\), to adjust the direction of the next update. The momentum is updated using \( g_{t+1} = \mu \cdot g_t + \frac{\nabla_xL(x_t, y)}{\|\nabla_xL(x_t, y)\|_1} \). The role of momentum is to maintain directionality throughout the optimization process and effectively circumvent falling into local optima. Finally, the new adversarial sample \(x_{t+1}\) is iteratively generated using \(x_{t+1} = x_t + \alpha \cdot \operatorname{sign} (g_{t+1}) \).

\subsubsection{TGR~\cite{zhang2023transferable}}

The TGR is an adversarial attack method specifically designed for Vision Transformers (ViTs). It enhances the effectiveness of attacks by reducing gradient variance during the training process. This method leverages the internal structural features of ViTs, diminishing the disparity in gradients among tokens, and thereby scaling the model's sensitivity to specific adversarial samples. Consequently, the adversarial samples generated are more likely to mislead different ViT models when transferred, inducing incorrect judgments. Notably, TGR demonstrates high efficacy and transferability in adversarial settings against various ViT and CNN models.

\subsection{Target Modification}
This category of methods exploits the characteristic of similarity among different models, such as attribution (explainability) similarity, by directly attacking these similar features to achieve transferable goals. Instead of directly attacking the model's cross-entropy, these methods often target the model's intermediate layers or attributions. Examples of this approach include Feature Importance-Aware Attack (FIA)~\cite{wang2021feature}, Neuron Attribution-based Attack (NAA)~\cite{zhang2022improving}, Double Adversarial Neuron Attribution Attack (DANAA)~\cite{jin2023danaa}, and Momentum Integrated Gradients (MIG)~\cite{ma2023transferable}.

\subsubsection{FIA~\cite{wang2021feature}}

The FIA achieves attack transferability by targeting key object-aware functions that significantly impact model decisions. Unlike traditional methods that indiscriminately distort features, leading to overfitting and limited transferability, FIA introduces an aggregated gradient approach. This method averages the gradients of a batch of randomly transformed versions of the image, emphasizing features related to the object and deemphasizing model-specific features. Such gradient information guides the generation of adversarial examples, aimed at disrupting key features, thereby enhancing transferability across different models.

\subsubsection{NAA~\cite{zhang2022improving}}
The NAA method first comprehensively attributes the model output to each neuron in the intermediate layer, then significantly reduces the computational cost through an approximation scheme. This scheme is based on two main assumptions: first, that the network's front half (feature extraction layer) and back half (decision layer) are independent in most traditional DNN models; second, that the gradient sequences of these two parts have zero covariance. As a result, NAA can make a faster and relatively accurate estimation of the importance of neurons. By weighting the attribution results of the neurons, it attacks the feature layer, thus generating transferable adversarial examples.

Specifically, NAA first uses the formula \(A_{yj} = \sum (x_i - x_i^\prime)\int_{0}^{1} \frac{\partial F}{\partial y_j} (y(x_\alpha ))\frac{\partial y_j}{\partial x_i} (x_\alpha )d\alpha \) to calculate neuron attribution. This formula measures the importance of neuron \(y_j\) by considering each input feature \(x_i\)'s effect on neuron \(y_j\) and neuron \(y_j\)'s contribution to the final output \(F\). Then, using the simplified computational assumptions, the attribution formula becomes \(A_{yj} \approx \Delta y_j \cdot IA(y_j)\). Here, \(IA(y_j)\) is the integrated attention, and this approximation allows for a rapid assessment of neuron importance. Finally, the target of perturbation generation is to minimize the weighted attribution \(W A_y = \sum_{A_{y_j} \geq 0} f_p(A_{y_j}) - \gamma \cdot \sum_{A_{y_j} \leq 0} f_n(-A_{y_j})\). This process adjusts the input image to reduce the model output's reliance on positive features while enhancing the impact of negative features, thereby improving the performance of transferable adversarial samples.

\subsubsection{DANAA~\cite{jin2023danaa}}

The DANAA method (Double Adversarial Neuron Attribution Attack) is an attack technique based on double adversarial neuron attribution. Its core principle lies in updating perturbations via a non-linear path, thereby more accurately assessing the importance of intermediate-layer neurons. The DANAA method attributes the model output to intermediate layer neurons, measuring the weight of each neuron and retaining features more crucial for transferability. This approach, by improving attribution results, enhances the transferability of adversarial attacks.

\subsubsection{MIG~\cite{ma2023transferable}}

MIG utilizes integrated gradient attributions to generate adversarial perturbations. Compared to traditional gradients, integrated gradients exhibit higher similarity across different models. MIG also incorporates a momentum strategy, optimizing the perturbation updates by accumulating integrated gradients from past iterations, thereby enhancing the attack's success rate and cross-model transferability.

Specifically, MIG starts by generating an initial zero perturbation \(\delta_0 = 0\). In each iteration, it calculates the gradient of the loss function of the current input image relative to the model \(\nabla_x L(f(x + \delta_t), y)\), and then combines this gradient with the momentum accumulated from previous iterations \(m_t\). The momentum update formula is \(m_{t+1} = \mu \cdot m_t + \frac{\nabla_x L(f(x + \delta_t), y)}{\|\nabla_x L(f(x + \delta_t), y)\|_1}\), where \(\mu\) is the momentum factor. The current perturbation is then updated using the accumulated momentum \(m_{t+1}\), following the formula \(\delta_{t+1} = \delta_t + \alpha \cdot \operatorname{sign}(m_{t+1})\), where \(\alpha\) is the step size. By iteratively repeating this process, MIG can gradually construct more transferable adversarial perturbations.

\subsection{Ensemble Approach}

This category of methods employs an approach where the attack process combines multiple models, using queries from multiple models to enhance transferability. However, these methods have certain limitations in real-world scenarios, as it is challenging to obtain multiple surrogate models in practical applications. Methods that fit this category include Model ensemble attacks~\cite{liu2016delving} and Stochastic Variance Reduced Ensemble Attack (SVRE)~\cite{xiong2022stochastic}.

\subsubsection{Model ensemble attacks}
Liu et al.~\cite{liu2016delving} proposed generating more transferable adversarial samples by ensembling multiple models. The core idea of this method is to optimize an ensemble of white-box models to generate adversarial samples capable of deceiving other black-box models. Specifically, given \(k\) white-box models with softmax outputs \(J_1\) to \(J_k\), an original image \(x\), and its true label \(y\), the ensemble method solves the following optimization problem:
\small
\[ \operatorname{argmin}_x -\log \left( \textstyle{\sum_{i=1}^{k}\alpha_i J_i (x)} \right) \cdot 1_y + \lambda  d (x,x^\prime)\]
Here, \(y\) is the target label specified by the attacker, \(\alpha_i J_i (x)\) represents the ensemble model, and \(\alpha_i\) are the ensemble weights (satisfying \(\textstyle{\sum_{i=1}^{k} \alpha = 1}\)). The goal of this optimization is to generate adversarial images that maintain their adversarial nature against an additional black-box model \(J_{k+1}\). The key to this method is that it is not only effective against a single model but can also cross different models, demonstrating strong transferability.

\subsubsection{SVRE~\cite{xiong2022stochastic}}
The SVRE operates on the principle of reducing gradient variance in model ensemble attacks to improve the transferability of adversarial samples. In traditional model ensemble attacks, attackers simply merge outputs from multiple models, but this approach neglects differences in gradient variance between models, possibly leading to local optima. SVRE reduces this variance through a two-level loop approach: the outer loop computes the average gradient of all models and passes the current sample to the inner loop; the inner loop performs multiple iterative updates, calculating the current gradient on a randomly selected model in each iteration, and adjusting it according to the gradient deviation in the outer loop. This method results in more accurate gradient updates in the outer loop, avoiding the issue of "overfitting" to the ensemble model and enhancing the transferability of adversarial samples to unknown models.

\section{Description of TAA-Bench}

\subsection{Algorithms Implementation}
In TAA-Bench, we consider 10 different types of adversarial attack methods as the current solution, including I-FGSM~\cite{kurakin2018adversarial}, DI-FGSM~\cite{xie2019improving}, MI-FGSM~\cite{dong2018boosting}, SI-NI-FGSM~\cite{lin2019nesterov}, NAA~\cite{zhang2022improving}, DANAA~\cite{jin2023danaa}, SSA~\cite{long2022frequency}, MIG~\cite{ma2023transferable}, AdvGAN~\cite{xiao2018generating}, and GE-AdvGAN~\cite{zhu2024geadvgan}. We select the methods as either classical or state-of-the-art approaches for TAA methods. 

Classical methods serve as baselines to measure the advancements of other newly improved algorithms. Moreover, these methods are chosen for their practical popularity and reproducibility in related studies, since some algorithms may involve a large number of hyperparameters leading to uncertainty in results and difficulties in implementation. In this case, our benchmark does not include such methods. The goal of TAA-Bench is to make the usage of TAA methods as simple and practical as possible to facilitate the in-depth analysis. We reflect the limitation of TAA-Bench by continuously including latest research results in our benchmark.

\subsection{Codebase of TAA-Bench}
We have constructed an extensible and modular codebase as the foundation for TAA-Bench. It consists of three modules: configuration, attack, and network model modules.

\textbf{The configuration module} comprises a YAML file for defining experimental parameters. This setup facilitates adaptable, reproducible experiments by clearly outlining variables such as network specifications and algorithm hyperparameters. Ensuring consistency and ease of modification, this module aligns with the imperative of reproducibility in scientific research.

\textbf{The attack module}, employing a modular architecture, encapsulates all the adversarial attack methods. The module aids future researchers for code review or extending new methods. Thus, the module provides a universal, dynamic toolkit to simulate and analyse the performance of transferable adversarial attack methods.

\textbf{The network model module} incorporates ten classic model structures, including Inception-v3, Inception-v4, ResNet-50, ResNet-101, ResNet-152, Inception-ResNet-v2, Inception-v3-adv, Inception-v3-ens3, Inception-v3-ens4, Inception-ResNet-v2-ens-adv in PyTorch. These models ensure that all attack methods can be thoroughly tested under the same structures, ensuring fairness in comparative experiments. Additionally, researchers can add specific model structures in this module for assessment and testing.

\section{Conclusion}

In summary, this paper provides an extensive review and benchmarking of the state-of-the-art techniques in the transferability of adversarial attacks, offering significant insights into the field of machine learning security. We have conducted a thorough analysis and categorization of a variety of methods. Our benchmarking efforts of TAA-Bench cover ten different adversarial attack methods, providing a comprehensive assessment of their effectiveness across various model architectures. In future work, we plan to continuously expand our benchmark and incorporate data analysis methodologies, such as interpretability analysis, to conduct an exhaustive evaluation of all replicated methods. This is anticipated to lead to new discoveries and further advancements in the field.






\bibliographystyle{IEEEtranS}
\bibliography{IEEEabrv,sample}
%



\end{document}